# Inferential Mechanics Part 1: Causal Mechanistic Theories of Machine Learning in Chemical Biology with Implications

Ilya Balabin and Thomas M. Kaiser[*]

*Avicenna Biosciences, Inc., Durham, NC, USA*

## Abstract

Machine learning techniques are now routinely encountered in research laboratories across the globe. Impressive progress has been made through ML/AI techniques with regards to large data set processing. This progress has increased the ability of the experimenter to digest data and make novel predictions regarding phenomena of interest. However, machine learning predictors generated from data sets taken from the natural sciences are often treated as "black boxes" which are used broadly and generally without detailed consideration of the causal structure of the data set of interest. Work has been attempted to bring causality into discussions of machine learning models of natural phenomena; however, a firm and unified theoretical treatment is lacking. This series of three papers explores the union of chemical theory, biological theory, probability theory and causality that will correct current causal flaws of machine learning in the natural sciences. This paper, Part 1 of the series, provides the formal framework of the foundational causal structure of phenomena in chemical biology and is extended to machine learning through the novel concept of focus, defined here as the ability of a machine learning algorithm to narrow down to a hidden underpinning mechanism in large data sets. Initial proof of these principles on a family of Akt inhibitors is also provided. The second paper containing Part 2 will provide a formal exploration of chemical similarity, and Part 3 will present extensive experimental evidence of how hidden causal structures weaken all machine learning in chemical biology. This series serves to establish for chemical biology a new kind of mathematical framework for modeling mechanisms in Nature without the need for the tools of reductionism: inferential mechanics.

## Section 1. Introduction

The study and practice of physics, chemistry and biology have seen significant changes with the increased use of artificial intelligence techniques like machine learning over the past decade.[1-6] Expansive data sets can now be automatically processed at breakneck speed, and even the process of data collection itself can be automated through the use of AI approaches.[7-10] Data integration across scientific domains has also seen tremendous benefit when using machine learning approaches for multi-omics investigation tools.[11-13] Major techniques like decision trees, random forests, neural nets, graph network models and deep learning have been applied to a wide range of chemical and biological systems with the goal of finding models capable of generalization across chemical and biological property prediction (e.g. target potency prediction or protein folding).[14-18] In the case of machine learning, generalization is the ability of a statistical model to fit rules which are consistent with available data and that will also apply to novel instances.[19] However, there have been increasingly frequent reports of problems with generalized accuracy and hallucination when machine learning is employed to model phenomena in the natural sciences.[20]

In order for a model to generalize and have high accuracy prospectively, the training and validation data used in ML construction must be generated by the same rules that will govern the production of future data.[19] Part of the response to these generalization problems has been to establish best practices for using machine learning alongside new methods of data curation as well as renewed discussions of applicability domains.[10, 21] Despite improvements in best practices, the work presented here will demonstrate that there are fundamental causal reasons for the increasing reports of underwhelming performance by machine learning methods in chemistry and biology, even in the cases where best practices are followed on high-quality data sets. We employ the mathematics of causality as it pertains to the statistical underpinnings of machine learning methods used in chemical biology to show formally that even high-quality data sets for properties like potency against a given target can have hidden rule changes within them. While there has been an initial report of evaluating causal machine learning in structure-activity relationships, we will demonstrate that this work is incomplete, and we offer a causal explanation for why test compounds that were more similar to the chosen seed structure for the Causal-Chemprop model as developed by Doyle *et al*. had a higher accuracy of prediction.[22] The concept of focus, the capacity for a machine learning training process to navigate hidden mechanistic structures in otherwise clean and correctly labeled data, will also be introduced. We will begin with a discussion of probability, causal calculus and why machines struggle when tasked with making predictions predicated on beliefs about Nature which are, unbeknownst to the scientist, untrue.

## Section 2. Results and Discussion

### *Section 2.1 – Introduction to Causal Calculus for Chemistry and Biology*

Machine learning is a field of computer science that enables machines to automatically detect patterns in data, and "the best way to make machines that can learn from data is to use the tools of probability theory."[23] Despite higher degrees of abstraction in chemistry and biology as compared to physics, there remains a fundamental physical link between the initial conditions and the resulting outcomes for each initial state.[24] To put it simply, there always exists a causal structure between the starting state of an experiment and the data obtained from that experiment.

It must be stated that the causal structure is not necessarily what the experimenter believes, even in the case of a well-controlled experiment. Chemical and biological scientists understand that there are often experimental errors which provide confounding reasons for why a given effect was observed (e.g. a small molecule sample was stored improperly and so the lack of activity against a target of interest was later discovered because there was no ligand actually present in the solution). However, there are other confounders as well. The scientist may believe something about the mechanism mediating a recorded data set, but that belief may not have any physical basis. As an example, data sets regarding a target protein structure-activity relationships are assembled for all compounds which have activities recorded for protein of interest; however, the actual allosteric or orthosteric mechanisms mediating those structure-activity relationships are not included in these target activity data sets.[25] These hidden mechanistic details will often have distinct structure-activity rules despite sharing the same protein target,[26-29] and these hidden mechanistic links may result in poor machine learning models when algorithms are trained on that otherwise high-quality target activity data set. To fully understand why this occurs, we need to explore causal calculus and the nature of causality as first formalized by Judea Pearl.[30] We will begin with a brief introduction to the formalisms of causal calculus including a causal model, a directed acyclic graph of the causal model, and the concept of a *do* operator. These are new formal concepts in chemical biology, and we wish to ensure that the causal graph for structure-activity relationships presented in Figure 4 can be understood by chemical and biological scientists. Additionally, an evaluation of Simpson's Paradox and the consequences of causality and causal inference for chemical biology and machine learning will become fully apparent. These are outlined in Table 1, Table 2 and Figure 3. If the reader is already familiar with causal calculus, we recommend skipping to *Section 2.3 – Causal Calculus for Structure Activity Relationships.*

*Section 2.2 – Causal Calculus Formalisms*

There are important semantic definitions required to understand the logic and framework of Causal Calculus. Importantly, a model in the context of causal calculus is a mathematical object that assigns truth values to sentences constructed in a spoken language. A causal model, *C*, is a triple of the following elements:[a,30]

$$C = \langle U, V, F \rangle$$

*where:*

*U is a set of variables intrinsic to the background universe which are unobserved and not determined by the model. These are our confounding variables*

*V is a set of observable variables that are determined within the model*

*F is a set of functions such that each $f_i$ maps the set of background variables, U, to the set of endogenous variables V. Functions that map between subsets of V are also permissible*

Furthermore, every causal model can be associated with a directed acyclic graph, *G(C)*, where directed edges point from objects in sets *U* and *V*, and the flow from the initial events in the graph is directed towards at least one outcome variable of interest. Each edge indicates the presence of a function in function set, *F*, which is the mathematical link between variables drawn from *U* or *V*. An example directed acyclic graph is given in Figure 1, and the acyclic nature of the graph refers to the directionality of the arrows representing *F* not allowing a totally clockwise or counterclockwise motion around the graph.

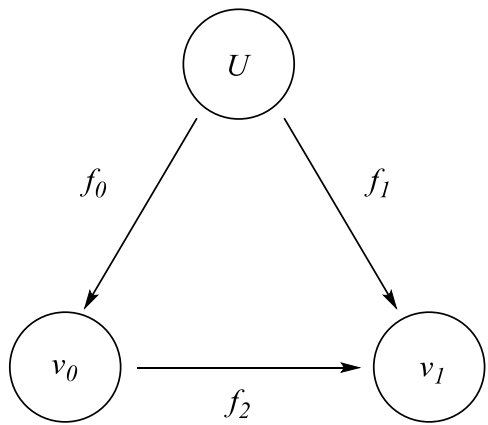

Figure 1. An example of a directed acyclic graph of a causal model, *G(C)*.

---

[a] *C* is used as the variable for the causal model here instead of the standard *M*. This is because *M* is used at a later point in the manuscript.

The utility of this framework arises as the graph provides a means to evaluate causal statements, confounding variables and likelihood distributions for causal statements. As an example from figure 1, $f_2$ is the function that maps the causal variable $v_0$ to the effect outcome $v_1$. Both $v_0$ and $v_1$ belong to the observable set V. Additionally, there is a confounding relationship between $v_0$ and $v_1$ that exists through the variable set, $U$, containing unobserved variables. The function set, $F$, is comprised of mathematical statements revolving around the *do* operator: what happens in the presence or absence of an action, in other words.[30] Formally, we can write $f_2$ in Figure 1 as follows:

$$f_2 = P(v_1 | do(v_0))$$

The *do* operator is a mathematical object with a high degree of utility in that it allows for the construction of probability statements regarding the likelihood of outcomes conditioned on the presence or absence of an action.[30] However, it is often very difficult to construct *do* statements as the observer requires both the action to be taken and untaken for the *do* operator to be developed. To use the terms of causal calculus, both the action's result and the inaction's result, the counterfactual, are required. As a consequence, counterfactual inference cannot be easily formalized in the standard methods of probability equations and this is especially true when variables of interest are unobserved in observational data sets.[30] Causal calculus affords us some powerful solutions to these problems, fortunately. Backdoor and Front door adjustments can be used to evaluate *do* operator statements to determine identity equations where the *do* operator is transformed into relationships which are calculable given the causal graph and some data set of interest.[31, 32] With the rules of identity equation determination, we can evaluate causal likelihoods even when key variables of interest are accounted for in a causal graph but not directly observed in the data. An excellent review alongside Pearl's original textbook on Causality are cited here so that those not familiar with these methods may have a primer.[30, 33]

The final conceptual framework needed to appreciate the statistical problem is the difference in the Total Effect (TE) versus Direct Effect (DE) in causal statements.[30] Total Effects are those probability statements that evaluate the total effect of an intervention variable on a response variable. In the example above in Figure 1, the intervention variable would be $v_0$ and the response variable would be $v_1$. However, models in the natural sciences do not seek to measure the Total Effect of the variable of $v_0$ on $v_1$ as $U$ is varied but instead to measure the Direct Effect of $v_0$ on $v_1$ with all other variables in the causal graph held constant. This is the Direct Effect of $v_0$ on $v_1$ and is represented by Figure 2.

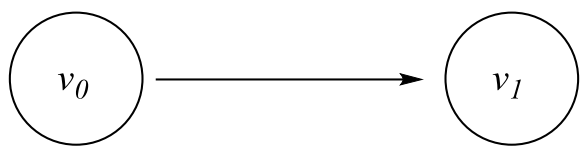

Figure 2. A directed acyclic graph of the Direct Effect of $v_0$ on $v_1$ with $U$ held constant.

### *Section 2.3 – Causal Calculus for Structure Activity Relationships*

The introduction to causality in Section 2.2 provides the tools needed to describe machine learning in chemical biology within the formal language of causality. At the heart of the study of chemical biology is the generation of an understanding for how a chemical space of interest modulates biological objects under study. The chemical stimuli represented by the set of structures, *S,* and the corresponding biological effect represented by the set of activities, *A*, are the recorded variables, and this Direct Effect of Structure-Activity relationship is what is sought when building machine learning models.

However, the desired Direct Effect Structure-Activity relationship in the data set for a protein target of interest can be masked by confounding variables such as structure-solubility relationships, experimental record error, and so on.[34-41] Accordingly, there have been many attempts to identify commonly encountered confounders with SAR data generation (e.g. the report of Pan-assay Interference Compounds or PAINs).[42] Beyond controlling for solubility and other standard confounders in the background set, *U*, in dose-response curves, we will now show that there is an additional causal variable which is a critical aspect of Structure-Activity relationships: the set of binding dynamics for each mode of action, *M*, where a small molecule binds to a protein and how the molecule exerts its effect. This site or mode of action is a crucial part of the understanding of how structure impacts function; and, this aspect often goes unappreciated, or at least unacknowledged, in machine learning with ligand-based methods as evidenced by the omission of protein surface mechanism tags in contemporary biological data sets.

This is where our problem for machine learning in chemistry starts to appear. Importantly, structure-activity trends taken on average for a single target may be reversed when subgroups of structures in *S* are split out according to *M;* this reversal in trend when a group of data are split according to subclasses and reanalyzed is known as Simpson's paradox.[43, 44] Compounds are grouped according to assay or target in contemporary cheminformatics databases, and the atomistic mechanistic details are not present in these public databases. We will see how this seemingly paradoxical reversal hinders our ability to generate machine learning models with the following thought experiment.

In this thought experiment, there exists a data set, *D*, for a protein of interest, protein X. This data set contains two medicinal chemistry series $s_1$ and $s_2$ with each molecule's structure and $IC_{50}$ activity represented data in Table 1. These are the data that we would pull from a source like ChEMBL so that we could construct an activity model of protein X $IC_{50}$ using classification ligand-based methods. Here, the activity class for each molecule will be equal to 1 if the $IC_{50}$ is below a desired value (in this case < 100 nM), and we can see this populated accordingly in the Activity Class column in Table 1. When the value is equal to 1, the compound has an $IC_{50}$ value below 100 nM whereas the value in the Activity Class column is 0 when the $IC_{50}$ is greater than or equal to 100 nM.

| Entry | Structure | $IC_{50}$ (nM) | Activity Class (<100 nM = 1) | Scaffold | Entry | Structure | $IC_{50}$ (nM) | Activity Class (<100 nM = 1) | Scaffold |
|---|---|---|---|---|---|---|---|---|---|
| 1 | 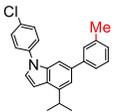 | 95 | 1 | $S_1$ | 5 | 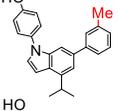 | 82 | 1 | $S_2$ |
| 2 | 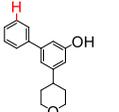 | 253 | 0 | $S_1$ | 6 | 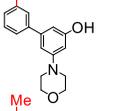 | 386 | 0 | $S_2$ |
| 3 | 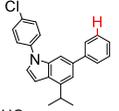 | 67 | 1 | $S_1$ | 7 | 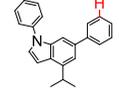 | 62 | 1 | $S_2$ |
| 4 | 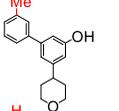 | 225 | 0 | $S_1$ | 8 | 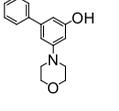 | 174 | 0 | $S_2$ |

Table 1. Hypothetical Data set, *D*, retrieved from a public database for Protein X.

Fingerprinting is a common means of representing chemical structure in a machine-readable format.[45, 46] Beyond the well-established nature of fingerprint representation in cheminformatics, there is the additional benefit that chemical features are preserved with high fidelity using these methods. To use a concept from information theory, the mutual information between the chemical structure and the fingerprint is much higher than the mutual information between the chemical structure and some set of calculated descriptors like tPSA, rotatable bonds, and cLogD.[47] Our goal when employing machine learning is to model the fundamental links between the mosaic of a molecule's functionality and the resulting biological response when a biological system is modulated by the molecule, and fingerprinting is accordingly a good method to parse each molecular entity in a machine readable format.

In our case, we are focused on the presence or absence of an aromatic methyl group in each of the Protein X inhibitors (indicated by the red color in Table 1). We can see the fingerprint bit encoding aryl-methyl as well as aryl-H in Table 2 where each chemical entry has been transformed into a machine-readable bit array and we are "zoomed in" on the aryl-methyl region of this bit vector. All the machine can see is the correlation between the fingerprints and the activity class. If the entirety of the cleaned and high-quality data set containing both $s_1$ and $s_2$ is used for machine learning as is current best practice, the Ar-Me fingerprint bit as an aspect for modelling potency by the ML algorithm will be ignored as there is maximum information entropy (i.e. maximum "surprise") for this bit.[48] In other words, using the presumptions of seeking the Direct Effect of Structure on Activity with the way we have constructed the data set, $D$, it appears that the presence of the Ar-Me on active compounds is randomly associated with activity or inactivity in this data set.

| Entry | Fingerprint Bits | | Activity Class (<100 nM = 1) |
|---|---|---|---|
| | ..., ArC-Me | ArC-H, ... | |
| 1 | 1 | 1 | 1 |
| 2 | 0 | 1 | 0 |
| 3 | 1 | 1 | 1 |
| 4 | 0 | 1 | 0 |
| 5 | 0 | 1 | 1 |
| 6 | 1 | 1 | 0 |
| 7 | 0 | 1 | 1 |
| 8 | 1 | 1 | 0 |

Table 2. Fingerprinted Data set, $D$, for all compounds with potency data for Protein X.

To see how this is problematic, we imagine a team at a later point conducts a structural chemical biology investigation on Target X and it is discovered that there are two binding sites on Protein X labeled $m_1$ and $m_2$ as seen in Figure 3. It is also determined that molecular series 1, $s_1$, exerts its enzyme inhibition activity at site $m_1$ and molecular series 2, $s_2$, exerts its enzyme inhibition activity at site $m_2$. The chemical and biological basis for structure-activity relationships that govern $m_1$ will be different from $m_2$ if the protein surfaces are distinct from each other. It is here that we see Simpson's paradox manifest. If we actually need to group compounds not by protein target (as is currently considered correct) but instead by binding pocket and the dynamics of binding, machine learning as it is currently practiced will miss that aryl-Me needs to be present for compounds in the $s_1$ family which hit $m_1$ and the aryl-Me group needs to be absent for the $s_2$ series exerting their activity at $m_2$. In our example, we can see that grouping compounds by $M$ reveals that the aryl-Me is essential for activity in *m₁* whereas the opposite trend holds for *m₂*! Although this is an idealized case to highlight the power of Simpson's paradox to limit machine learning model generation for a combination of scaffolds on a single target, there are many examples of cellular data where EC$_{50}$ data values are grouped together by assay type (e.g. Respiratory syncytial virus replication inhibition data in HepG2 cells) where multiple scaffolds are grouped together in the activity data set and yet many different proteins or even RNA/DNA mechanisms are involved. Frequently, the mechanism is not known or reported for collections of scaffolds when cellular activity data are reported. As can be understood from principles drawn from chemistry and pharmacology,[26-29] distinct binding pockets on the same protein have little overlap in their SAR let alone the similarity in SAR trends between binding pockets of unrelated proteins. This is problematic if we seek the ability to model the SAR trends of *s₁* or *s₂*.

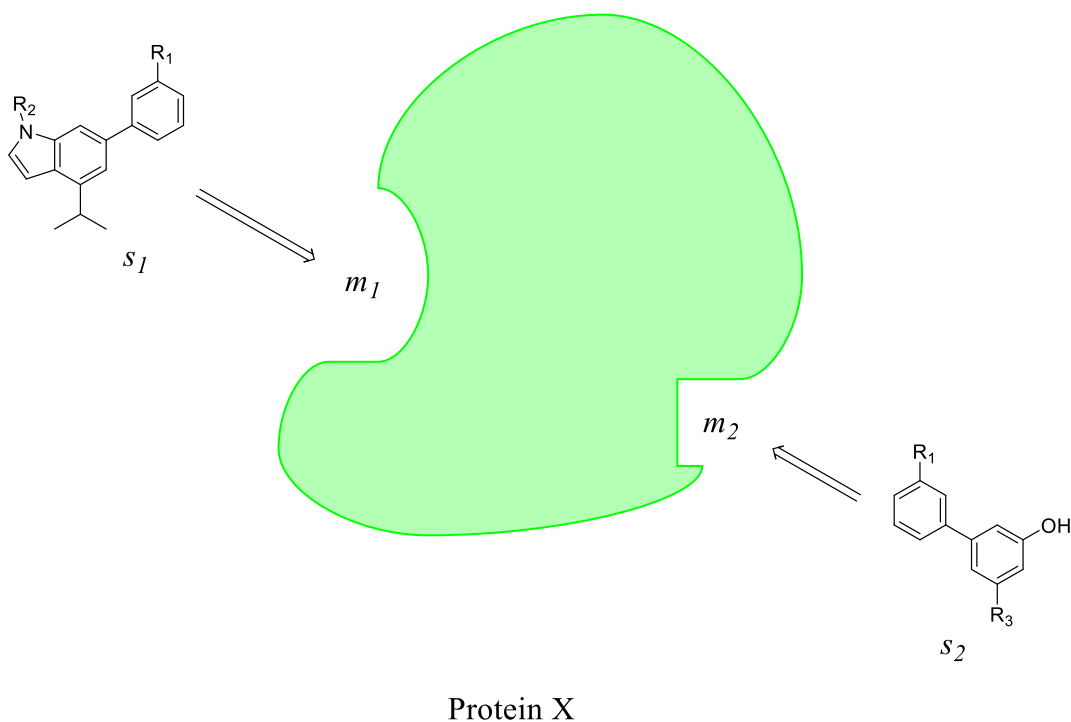

Protein X

Figure 3. Molecular interaction basis for data set, *D*, consisting of activity data for *s₁* and *s₂* on protein target X.

With this result of our thought experiment, we can see that the Direct Effect of chemical structure on biological activity is only measurable if the Mediator variable set, $M$, representing the dynamics of binding a small molecule to a target is held constant alongside the set of variables in the universe not determined by the model, $U$. Therefore, the chemical structure and biological activity relationship can be represented by a graph as drawn in Figure 4:

*where:*

*$U$ is a set of variables intrinsic to the background universe which are unobserved and not determined by the model*

*$S$ is a set of chemical structures and their associated fingerprints under investigation in a biological assay*

*$M$ is a set of small molecule-protein complexes which mediate the influence of $S$*

*$A$ is a set of observable biological activities resulting from the formation of small molecule-protein complexes within the assay environment*

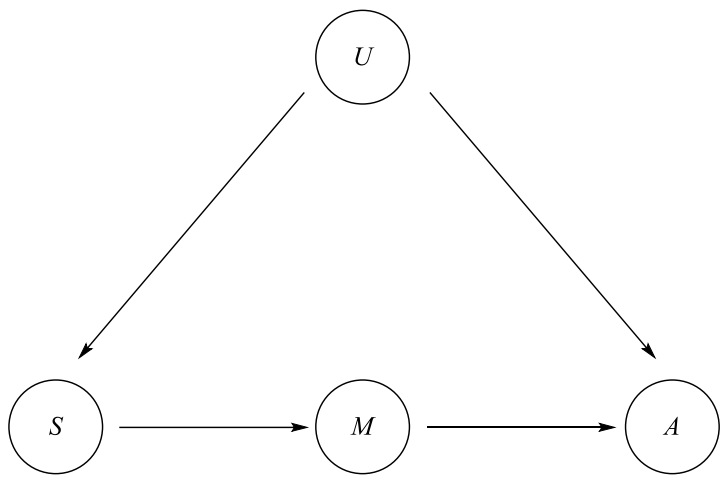

Figure 4. The directed acyclic graph of the idealized causal structure-activity relationship, $G(C)$.

We can now appreciate the problem for modelling this graph with any statistical method including automated statistical modelling conducted by machine learning. We want the Direct Effect of Structure on Activity but we instead have a data set with the Total Effect where chemical structures, $S$, exert their effect on Activity, $A$, through a variable mechanistic mediator set, $M$. Therefore, we need a way to condition on the mediator variable, $M$, so that we can evaluate the

Direct Effect of Structure on Activity. The Front Door Adjustment method allows for us to block any confounding associations mediated by unobserved variable set, U, so that we can evaluate the Causal path from S to A and solve for the identity equation of the causal structure-activity probability - P(a|do(s)). The reason for employing the Front Door Adjustment for the graph in Figure 4 is that the set U, containing unobserved universal confounding variables like solubility, is unobserved and we are therefore unable to condition our *do* operator statements on U.[30] The Front Door Adjustment can be done by identifying the causal effect of S on M as well as the causal effect of M on A.

We seek a calculable probability expression which will allow us to account for M and let us evaluate the causal probability equation between S and A. Accordingly, it will be demonstrated that it is not possible to determine an identity equation for the probability statement without knowing the distribution of M.

*We are in pursuit of identification equation for the causal probability between S and A:*

$$P(a|do(s))$$

*According to the rules of do calculus and our causal graph, we know that the desired probability statement is equal to the following:*

$$P(a|do(s)) = \sum_m P(m|do(s))P(a|do(m))$$

*Where the total causal relationship is a sum of the distribution of all values of M.*

*The causal effect of s on m is simply:*

$$P(m|do(s)) = P(m|s)$$

*Additionally, there is a backdoor path between M and A, so we will need to condition on S when we transform the do operator for P(a|do(m)) according to the following adjustment relationship to evaluate the causal effect of m on a over the distribution of s:*

$$P(a|do(m)) = \sum_s P(a|m,s)P(s)$$

*Therefore, the Front Door Adjustment criterion allows us to construct the identification equation as follows:*

$$P(a|do(s)) = \sum_m P(m|s) \sum_s P(a|m,s)P(s)$$

We now have an equation that lets us calculate causal likelihoods of *S* on *A* given the intermediate *M*. However, a significant problem has revealed itself: we need to know the distribution of *M* in order to use this identity equation! What are we to do when the binding site mechanisms, the mediators of Structure-Activity relationships, are unknown?

Given Simpson's paradox, we must find a way to partition out dissimilar modes of action so that we can construct a machine learning model of our assay or property of interest. Until this point, our only recourse would have been to reductionistically explore all molecules and categorize them according to *M* so that we could then build an ML model on only those structures that exerted their biological effects in a similar way. We could imagine determining crystal structures for each series in the data set when bound to a target of interest or we could employ mutagenesis on the protein to determine mechanism of binding for each scaffold, and then we could group the structures into data sets according binding mode for a target. This is prohibitively expensive for data sets of even just a few hundred compounds let alone data sets of tens of thousands of compounds.

### Section 2.4 – Implications for Simpson's Paradox in Machine Learning Models of Structure-Activity Relationships with an Unknown Distribution of M

The key insight to solving this problem inferentially without reductionistic tools is to realize that we can group molecules by chemical series where we assume that each series will have a constant mechanism for its properties of interest that governs that series (even in the case that the mechanism is not known to the scientist). If we return to our thought experiment, $s_1$ works via $m_1$ and $s_2$ works via $m_2$. In effect, each scaffold series is mapping out an inverse map of their respective activity sites, as that binding pocket has a definite albeit unknown structure. We can use the fact that compounds that look very similar to $s_1$ will work like $s_1$ as all compounds that fit in the same pocket must do so according to the same rules as $s_1$. In other words, form is function in chemical biology.

To find the optimum training set within an otherwise high-quality contemporary data set for a protein of interest, we can select a single scaffold as a starting point for ML training and testing. We will use that single scaffold as the only source for the test set, and we will hold *M* constant for the test set as a consequence. Then, as we add the next most similar series from our data set to our starting scaffold for training, we can monitor the retrospective accuracy for testing on our initial scaffold as a function of increasing dissimilarity. In our case, we employ the ROC AUC as the retrospective accuracy measure for classification tasks. As we scan through the addition of increasingly dissimilar compounds, we can look for signs that additional high quality assay data for the same target are actually degrading our training for that constant mechanism in the test set for the initial scaffold. To control for *M* by starting with a single scaffold and then scanning away from that scaffold in training using similarity affords us a new concept in machine learning: focus. Focus is the ability to inferentially home in on the real causal relationship between structure and function, even when *M* is unknown, and to establish the structural boundary of the validity of that relationship. If we are correct, we expect to see the retrospective accuracy increase and peak for some set of similar compounds. Then, as we add more compounds believed to have the same mechanism because they hit the same target, we will see a decay in accuracy as we unwittingly cross into a new *m* that is distinct from our original series.

Formally, the process for inferentially drawing the cutoff at the correct mechanistic subset, $\Delta_s$, for a selected series of compounds, *s*, within a protein target activity set of interest, *D*, is as follows:

For $D = \{x_i\}_{i=1}^{N}$ to be used for machine learning

> *where $x_i$ is the ith molecule in some data set D*
>
> *and there are N molecules in D.*
>
> *Where we select a decision value threshold T = (0, ∞) nanomolar*
>
> $y_i = 1$ *when* $IC_{50,x_i} < T$;
>
> *if this condition for active is not met, $y_i = 0$.*
>
> *Where N is converted to a variable, $N_{sim}$, the number of increasingly dissimilar included molecules relative to s, a single uniform series of interest from the literature data set, D.*
>
> *Similarity is defined using a similarity metric in relation to s.*
>
> *$N_{sim} = [0, n]$, the size of the addition to the training set where n = the number of remaining molecules in D which are not in s.*
>
> *We will select the optimum $N_{sim}$ and T via testing on a reserved 50% of s and evaluating the retrospective accuracy of a machine learning predictor trained on the complementary 50% of s + $N_{sim}$. Evaluation of accuracy will be through ROC AUC averaged across multiple evaluations performed on random selections of 50% of s for testing.*
>
> *This creates the data subset $\Delta_s = s + \{x_i, y_i\}^{N_{sim}}$ consisting of compounds belonging to s plus the optimum $N_{sim}$ at the determined T. This is the mechanistic set to be used for machine learning modeling of a property of interest for a prespecified scaffold, s.*

The reason for evaluating the classification decision threshold, $T$, is that some decision values will have high information entropy due to thermal noise or experimental error. We must find a classification threshold for our potency that is low enough to be meaningfully discriminatory regarding active and inactive compounds, but not so low as to be unreliably noisy. Care must be taken to balance the focus of a model alongside its fitness, and we lay out that balance as follows. By combinatorially exploring $T$ and $N_{sim}$ using a 50/50 train/test method for scaffold $s$ and evaluating the retrospective performance of each combination and averaging the performance in replicate (e.g. for 100 unique test-train splits at a given $T$ and $N_{sim}$ combination), we will be able to find the values of $T$ and $N_{sim}$ which generate the optimal learned function for modeling the Structure-Activity relationship of $s$ and a constant $m$. This high number of replications will ensure that a good or bad retrospective prediction result for some combination of $T$ and $N_{sim}$ is not a spuriously overfit model but instead correctly focused on $m$ for $s$. This process effectively allows us to use the fundamental properties regarding probability and inference to compete different mechanistic beliefs about the behavior of sets of similar compounds. Once we add just enough dissimilar compounds that finally cross into a distinct and separate mechanism, $m$, with respect to our starting scaffold, $s$, our ability to model reality inferentially incurs damage.

In Figure 5, we see an example of this expected machine learning behavior using a random forest classification algorithm being tested on a single scaffold of Akt kinase inhibitors taken from ChEMBL (300 compounds from the series first reported by Kettle et al.;[49] full details of experiment are available at https://github.com/balabin/IM_01). The 300 compounds were randomly split into 50% test and 50% train sets in an unbalanced fashion 500 times. For each of the 500 test sets, the corresponding training set was increased according to $N_{sim}$ with the average Tanimoto similarity to the scaffold training set used to sort the complementary Akt activity data set in decreasing similarity. We can observe the average ROC AUC for the test set comprised of 50% of this scaffold peaks at 0.841 for $N_{sim}$ = 25 most similar compounds when $T$ = 10 nM, and the average ROC AUC begins to decay as we move away into less similar chemical space. The ROC AUC continues to decline to 0.791 as we scan out to the entirety of the remaining Akt $IC_{50}$ data set at $N_{sim}$ = 2225. Despite all of these 2225 compounds all having high quality Akt activity data, only the most similar 25 are needed to give the best performing algorithm on the test set. Any additional data surprisingly degrades the learning as Simpson's paradox reveals a hidden mechanistic incongruency in the total structure-activity relationship for the entire Akt data set as compared to the mechanistic rules for our test set of scaffold $s$. The causal explanation for why test compounds that were more similar to the chosen seed structure for the Causal-Chemprop model had a higher accuracy of prediction is that those similar compounds likely shared a common mechanism, $m$, for activity against either Aurora Kinase A or Abl kinase.[22] The authors of the Causal-Chemprop paper posited that this observed behavior was due to the fact that the ground truth causal graph was not available, and that future work was needed to develop a more causal molecular representation. In light of the present work, we can now see that the mediating mechanism distribution for *M* was the unknown aspect, and we can make use of the ground truth causal graph for structure-activity relationships presented in Figure 4 to navigate the hidden distribution of *M* inferentially via focus.

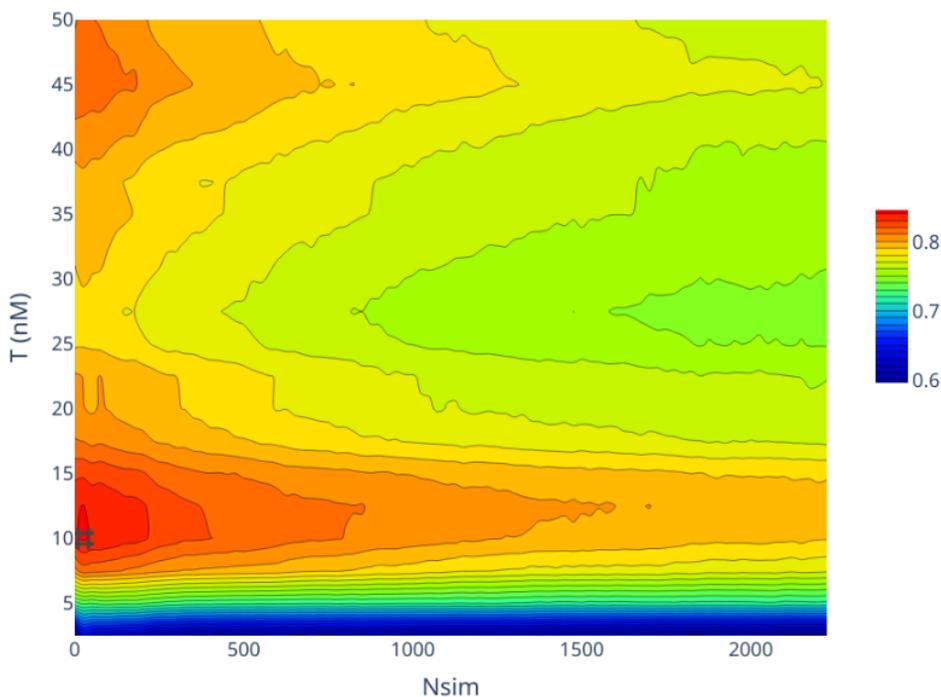

Figure 5. The behavior of the retrospective performance of a classification algorithm on a test set of *s* as a function of training set size when the training set is sorted by similarity to *s*. Average ROC AUC is recorded as a color according to the legend on the right of the figure. The hash mark indicates the ROC AUC maximum at *T* = 10 nM and $N_{sim}$ = 25.

It is important to note that the incongruency that is expected to materialize will only be observed if at least one of two things occurs. The first is there must be at least one additional *m* in the data set. Some proteins will only have one mode of action in their data sets, and we would expect $N_{sim}$ to equal the maximum size, *N*, of the data set in this case. The second possibility is that there is experimental error within the new Structure-Activity data of an adjacent series that causes a severe incongruity in the SAR patterns rather than an incongruity as consequence of a mechanistic change. In this later case, we would argue that it is imperative to ignore this erroneous data as well. It is only possible to infer mechanistic distinctions when the data sets for a property or phenomenon of interest are of high quality and well controlled; inferentialism is not a replacement for a well-controlled experiment. Fortunately, resources like ChEMBL ensure that well controlled data are contained in most data sets that can be downloaded.

As is often the case, new answers bring new questions. This work treats similarity as an abstract idealized concept but inferential mechanics requires a sound definition of similarity. Part 2 of this series will formally explore similarity for pharmacophore space and establish a "yardstick" for pharmacophore space empirically. With Part 2's work as a prelude, Part 3 of this series will evaluate the notions of inferential mechanics in a test case where the high-quality data set for Akt $IC_{50}$ values is taken from ChEMBL and a variety of scaffolds are evaluated employing inferential mechanics. Additionally, prospective applications of inferential mechanics for RSV Fusion protein inhibitors, c-Abl inhibitors and ROCK1/2 inhibitors will be discussed.[50-52]

## 3. Conclusion

In this work we show that there are hidden mediators of structure-activity relationships within data sets of chemical biology regarding even single target data sets. In the simplest case, data sets regarding a single protein can frequently have orthosteric inhibitors and allosteric inhibitors lumped together in a single "activity" data set for a protein of interest. These hidden and distinct means of structure exerting activity create problems for modelling such data sets with techniques like machine learning as the underpinning rules for probability result in significant paradoxes when the causal structure of a data set is flawed. Historically, the only means of correcting this problem was to rely on the tools of reductionism like crystallographic binding site determination. Here, we propose that the very problems of information incongruity that give rise to Simpson's paradox can be used to detect when the addition of otherwise correctly labeled data is actually working via different and hidden mechanisms. When a uniform and carefully curated starting scaffold is selected as a starting point, errors in mechanistic belief manifest themselves in terms of reductions in retrospective accuracy as a data set is scanned through the use of similarity to said starting point. We define this principle here as focus. This work provides the theoretical foundation for explaining why scaling to very large training sets to increase generalizability is nonsensical; to obtain the maximum accuracy for specific regions of chemical space, the experimenter must focus the training set down to datapoints which are produced via the same mechanistic rules. The result of machine learning on that focused data set is an inferential mathematical rule not generated with the tools of reductionism, and yet this inferential rule is still descriptive of discrete mechanistic phenomena within our universe.

## 4. Acknowledgements

There are many people who have given their effort and criticism to this work. Without Prof. Dennis Liotta providing the resources of his laboratory a decade ago, these ideas would have never germinated. Thomas Kaiser is deeply indebted to Prof. Liotta and all of the members of his laboratory who worked on TMK's early machine learning programs. Drs. Pieter Burger and Zackery Dentmon were instrumental in generating the prospective proof of inferential mechanics over a lengthy collaboration, and none of this work would have been possible without them. Finally, Mrs. Michelle Kaiser has given her extraordinary patience and support to this work. Without her, these new theories would never have been realized.

Additionally, Prof. Russell Greiner, Dr. Homer Pearce, Dr. Sandro Belvedere, Prof. Casey Wade, Dr. Parker Dryja, Mr. Hyunjoon Kim and Dr. John Olivarius-McAllister provided essential commentary on concepts and/or the draft of this manuscript. The authors are grateful for their commentary.

Finally, this work would not have been finished without the investment of DCVC Bio, and the authors are especially grateful to Mr. Christopher Meldrum, Dr. John Hamer and Dr. Kiersten Stead.